\documentclass[10pt,twocolumn,letterpaper]{article}

\usepackage[pagenumbers]{wacv} 

\usepackage{algorithm}
\usepackage{makecell}
\usepackage{algorithmic}
\usepackage{xcolor}
\usepackage{multirow}
\usepackage[table]{xcolor}
\usepackage{caption}
\usepackage{booktabs}
\usepackage{array}
\usepackage{amsfonts}
\newcolumntype{C}[1]{>{\centering\arraybackslash}p{#1}}
\usepackage{enumitem}
\usepackage{subfiles}
\usepackage{threeparttable}
\usepackage{subcaption}
\usepackage{newfloat}
\usepackage{listings}
\usepackage{tabularx}

\AtBeginDocument{%
  \setlength\abovedisplayskip{6pt plus 2pt minus 2pt}
  \setlength\belowdisplayskip{6pt plus 2pt minus 2pt}
  \setlength\abovedisplayshortskip{0pt plus 2pt}
  \setlength\belowdisplayshortskip{4pt plus 2pt minus 2pt}
}

\definecolor{wacvblue}{rgb}{0.21,0.49,0.74}
\usepackage[pagebackref,breaklinks,colorlinks,allcolors=wacvblue]{hyperref}
\usepackage{tabularx}
\usepackage{makecell}
\usepackage{threeparttable}
\usepackage{booktabs}
\usepackage{graphicx}

\def\wacvPaperID{****} 
\def\confName{WACV}
\def\confYear{2026}

\title{LLM Augmented Intervenable Multimodal Adaptor for Post-operative Complication Prediction in Lung Cancer Surgery}

\author{Shubham Pandey$^{\dagger}$, Bhavin Jawade$^{\dagger}$, Srirangaraj Setlur$^{\dagger}$, Venu Govindaraju$^{\dagger}$, Kenneth Seastedt$^{{\dagger}{\star}}$ \\
$^{\dagger}$University at Buffalo, $^{\star}$Roswell Park Comprehensive Cancer Center \\
Buffalo, NY, USA \\
{\tt\small \{spandey8, bhavinja, setlur, govind\}@buffalo.edu, Kenneth.Seastedt@RoswellPark.org}
}

\begin{document}
\maketitle
\begin{abstract}
Postoperative complications remain a critical concern in clinical practice, adversely affecting patient outcomes and contributing to rising healthcare costs. We present MIRACLE, a deep learning architecture for prediction of risk of postoperative complications in lung cancer surgery by integrating preoperative clinical and radiological data. MIRACLE employs a hyperspherical embedding space fusion of heterogeneous inputs, enabling the extraction of robust, discriminative features from both structured clinical records and high-dimensional radiological images. To enhance transparency of prediction and clinical utility, we incorporate an interventional deep learning module in MIRACLE, that not only refines predictions but also provides interpretable and actionable insights, allowing domain experts to interactively adjust recommendations based on clinical expertise. We validate our approach on POC-L, a real-world dataset comprising 3,094 lung cancer patients who underwent surgery at Roswell Park Comprehensive Cancer Center. Our results demonstrate that MIRACLE outperforms various traditional machine learning models and contemporary large language models (LLM) variants alone, for personalized and explainable postoperative risk management. Our codebase is available at \url{https://github.com/KNITPhoenix/MIRACLE}.
\end{abstract}    
\vspace{-0.75cm}
\section{Introduction}
\label{sec:intro}
Lung cancer remains the leading cause of cancer related mortality worldwide and surgical resection provides a 2.5-fold survival advantage over non‐surgical management. However, surgery carries substantial perioperative risk, with postoperative morbidity exceeding 40\% and mortality surpassing 5\%, underscoring the need for accurate, individualized preoperative risk assessment \cite{ALTORKI2018915,SiegelCancerstatistics,PuenteMaestuChestEarly,ROSEN20161850}. Current risk models, including those developed by the Society of Thoracic Surgeons (STS) \cite{STScalc}, primarily rely on global pulmonary function measures such as FEV1 and DLCO. While clinically useful, these population level indicators fail to capture nuanced patient‐specific interactions between lung tissue heterogeneity, comorbidities and tumor biology \cite{KEARNEY1994753,KomiciKlaraFrailty,SalatiMicheleRisk,BURT201419,TongBetty2024}. Furthermore, such models do not incorporate imaging biomarkers such as emphysema distribution \cite{DAI2017824}, parenchymal texture \cite{Huang2016281} or tumor‐associated remodeling \cite{PMID28205188}, which are known to influence postoperative pulmonary outcomes. As a result, surgeons currently depend on population level statistics and subjective clinical judgment which lead to suboptimal decisions where patients who might have tolerated surgery are denied surgery, while others at high risk undergo procedures without adequate postoperative planning \cite{BRUNELLI2013e166S}. Radiomics provides a quantitative pathway for capturing lung heterogeneity and tumor microenvironment through high‐dimensional CT features \cite{Williams2021Radiomics}. When combined with clinical variables, radiomics improves postoperative risk stratification \cite{Huang2016281,10.1371/journal.pmed.1002711,NationalComprehensiveNonSmall}. However, most radiomic and ML approaches function as static black boxes offering predictions without transparency, clinical interpretability or mechanisms for physician intervention. Meanwhile, large language models enable clinically grounded explanations \cite{Brown2020GPT3,Alsentzer2019ClinicalBERT}, but remain disconnected from risk prediction workflows.

In this work, we introduce \textbf{MIRACLE} (Multi-modal Integrated Radiomics And Clinical Language-based Explanation), a unified deep learning framework that integrates: (i) structured preoperative clinical features, (ii) CT derived radiomic biomarkers and (iii) LLM generated, evidence-grounded natural language explanations to support risk prediction and interpretability. Our key contributions are: \textbf{(i) Unified Multimodal Risk Modeling:} Joint fusion of clinical data, radiomics and language-based explanations to enable accurate, interpretable and clinician-editable postoperative risk assessment, \textbf{(ii) Bayesian MLP Architecture:} Robust learning under small, imbalanced datasets with calibrated uncertainty estimation that mitigates overfitting and \textbf{(iii) Real-World Evaluation:} Training and validation on \textbf{POC-L} dataset, consisting of 3,094 lung cancer surgical patients treated between 2009–2023 at Roswell Park Comprehensive Cancer Center. We benchmark MIRACLE against thoracic surgeons (mean experience 17.75 years), standard ML models and multiple open-source LLMs, achieving superior performance across AUC and True Postive Rate (Sensitivity) at clinically relevant False Positive Rates (1 - Specificity). To our knowledge, this is the first method that jointly integrates preoperative clinical and radiological features with clinician intervenable, natural-language explanations, transforming risk assessment from a static black-box output into an interactive, transparent clinical decision-support system.
\section{Related Work}
\label{sec:relatedworks}
In this section, we review prior contributions across three key domains that inform the development of our proposed framework:

\textbf{Postoperative Complication Prediction.}
Early work in postoperative risk stratification relied primarily on statistical models and classical machine learning models applied to large clinical registries \cite{noncardiacpostoperative}. More recent studies demonstrated that ensemble learning methods such as Random Forest and Gradient Boosting can improve predictive performance, reporting AUC gains of 3-5\% over logistic regression for major surgical outcome prediction \cite{ExplainableML}. However, these approaches treat risk estimation as a purely data-driven classification task and provide limited interpretability, lacking mechanisms for clinician interaction or intervention. Current workflows therefore remain dependent on population-level thresholds and physician judgment without integrated computational decision support. To our knowledge, no prior framework unifies structured clinical data, radiological biomarkers and explainable language model, outputs into a single predictive pipeline.

\textbf{LLMs in Clinical Decision Support.}
LLMs have recently shown strong potential in clinical reasoning tasks, including summarization, diagnostic reasoning and report generation. GPT-3 demonstrated zero and few shot proficiency on medical QA and clinical note summarization tasks \cite{Brown2020GPT3}, while Med-PaLM achieved state of the art results on biomedical QA benchmarks \cite{Singhal2023MedPaLM}. In radiology, instruction-tuned LLMs have been evaluated for automated report generation, achieving performance comparable to practicing radiologists in early assessments. Similarly, domain-adapted transformer encoders such as ClinicalBERT have been applied to clinical concept extraction and outcome prediction from electronic health records \cite{Alsentzer2019ClinicalBERT}. Despite these advances, existing approaches treat text generation and clinical outcome prediction as independent processes. No prior LLM based framework directly predicts postoperative complication risk in lung cancer surgery or integrates language based explanations jointly with multimodal clinical and radiomic predictors, to our knowledge.

\textbf{Explainability and Human--AI Intervention.}
Explainability is central to adopting AI for clinical decision making. Previous feature attribution methods such as LIME \cite{Ribeiro2016WhySI} and SHAP \cite{Lundberg2017AUA} provide insights into feature importance for clinical prediction models, including radiomics. More interpretable architectures such as Concept Bottleneck Models (CBMs) decompose predictions into clinician readable intermediate concepts \cite{Koh2020ConceptBM} but require extensive manual concept definition and annotation. Recent work has explored incorporating clinician feedback into AI inference pipelines \cite{Pang2024IntegratingCK}. However, these systems typically rely on predefined concept spaces or offer limited integration between explanation and intervention \cite{Hager2023BestOB}. Few existing frameworks unify high predictive performance, natural language explanations and clinician editable intervention mechanisms. MIRACLE addresses this gap by embedding LLM generated and evidence grounded explanations directly within a multimodal predictive architecture that supports real time clinician adjustment of model outputs.
\section{Methods}
\label{sec:methods}
In this section, we describe our proposed deep learning architecture for predicting risk of postoperative complications for lung cancer surgery candidates. Our method integrates preoperative clinical and radiological features with explainable remarks generated by a large language model (LLM). The key components of our approach are defined in the following subsections.

\subsection{Problem Definition}
Let $\mathbf{c} \in \mathbb{R}^{17}$ represent a vector of 17 preoperative clinical features (detailed in supplementary material) in our dataset. Due to the limitations of no direct image access from the collaborating cancer research center, radiological features were extracted from a patient’s chest CT scan using Total Segmentator \cite{Wasserthal_2023} and pyradiomics \cite{Pyradiomics} library, quantifying imaging biomarkers relevant for lung tissue characterization, represented by a vector of 113 preoperative radiological features (detailed in supplementary material) and denoted by $\mathbf{r} \in \mathbb{R}^{113}$. $M$ represent textual remarks generated by an LLM based on a clinical summary constructed from $\mathbf{c}$, a carefully crafted contextual knowledge bank $K$ and a predefined prompt $P$. The clinical summary is obtained via a deterministic function $S=f_{\text{summary}}(\mathbf{c})$, generating the remarks as:
\[
M = \text{LLM}(S, K, P)
\]
Our objective is to learn a function as follows:
\[
f:(c,r,M)\rightarrow\\y\ \in {0,1}
\]
which predicts the probability of postoperative complications \(y\). For this study, we consider a set of ten postoperative complication events (defined in supplementary material) \(\mathcal{C} = \{C_1, C_2, \dots, C_{10}\}\) and define
\[
y = 
\begin{cases}
1, & \exists\, C_i \in \mathcal{C}\,\text{such that patient experiences }C_i,\\
0, & \text{otherwise}.
\end{cases}
\]

\subsection{MIRACLE}
MIRACLE stands for Multi-modal Integrated Radiomics And Clinical Language based Explanation. The proposed architecture consists of three main modules:
(i) Two separate Bayesian MLP networks, one each for clinical and radiological features, 
(ii) An encoding module using a frozen encoder, fine-tuned on medical data for textual remarks and 
(iii) A fusion network for final prediction.
The processing pipeline is illustrated as follows:

\textbf{(i) Clinical Summary and Remarks Generation:} A fixed template first renders the raw clinical feature vector $\mathbf{c}$ into a coherent, human readable summary $S = f_{\text{summary}}(\mathbf{c})$. This summary enumerates each feature given in clinical vector, such as patient's age, smoking history, etc. This ensures strict traceability between structured inputs and generated text. This summary is concatenated with a domain knowledge bank $K$ and a task-specific prompt $P$ to form the composite input $\{S,K,P\}$ to a frozen LLM, yielding a natural language clinical remark $M$. Each remark $M$ highlights patient-specific risk or protective indicators that are grounded in both the patient’s clinical data and evidence based guidelines encoded within $K$, forming the interpretable explanation channel for the prediction model.

\textbf{(ii) Feature Embedding:} 
\begin{align*}
    E_c = f_{c}(\mathbf{c}) \in \mathbb{R}^{d}; E_r = f_{r}(\mathbf{r}) \in \mathbb{R}^{d}; E_m = f_{m}(M) \in \mathbb{R}^{d}
\end{align*}
where $f_{c}(\cdot)$ and $f_{r}(\cdot)$ are Bayesian MLPs based networks that encode preoperative clinical and radiological features respectively into a $\mathbf{d}$ dimensional vector and $f_{m}(\cdot)$ is a frozen text encoder, fine-tuned on clinical texts, that embeds the LLM generated remarks (also in a $\mathbf{d}$ dimensional vector) so that it can be used by the fusion network.

\textbf{(iii) Feature Fusion and Prediction:} The embeddings from each module is fused using a weighted sum, to form a global representation:
\[
E = (w_c*E_c + w_r*E_r + w_m*E_m) \in \mathbb{R}^{d}
\]
This fused embedding is then passed through a final Bayesian MLP Network $f_{\text{final}}(\cdot)$, with the output activated by a sigmoid function to yield the predicted probability of post-operative complications:
\[
\hat{y} = \sigma\left(f_{\text{final}}(E)\right),
\]
where $\sigma(\cdot)$ denotes the sigmoid function.


\subsubsection{Remarks Generation Approach:}
This component provides the explainability layer of MIRACLE, so surgeons can plan ahead of the surgical procedure and enables clinician guided intervention during inference. After deriving the clinical summary $S$ from the patient’s clinical data $\mathbf{c}$ using a predefined fixed template, it is concatenated with a predefined prompt $P$ and a knowledge bank $K$. The knowledge bank $K$ is a manually curated textual repository developed in collaboration with thoracic surgeons and oncology specialists. It encodes domain relationships linking preoperative clinical factors to postoperative complication risks across pulmonary and respiratory outcomes. Content is derived from authoritative textbooks, peer-reviewed surgical oncology literature and published peri-operative risk assessment guidelines. The knowledge bank includes descriptions of feature ranges, documented interactions between risk factors and mechanistic explanations of surgical complication pathways. We released knowledge bank alongside the codebase (link in abstract). The composite input $\{S,K,P\}$ supplied to an LLM directs the model to generate grounded, evidence-consistent explanatory remarks focused exclusively on interpretable clinical reasoning. We evaluated multiple LLM paradigms, including instruction-following, reasoning-oriented and domain-fine-tuned models, selecting the optimal performer for each experimental configuration based on clinical coherence and alignment with domain knowledge. Both the selected LLM and the downstream text encoder remain frozen throughout training and inference. This prevents hallucination drift, maintains linguistic stability of the explanation channel and preserves the clinical priors encoded during pre-training.

\subsubsection{Training Strategy:}
For training, we adopted focal loss \cite{lin2018focallossdenseobject}, as it was built to address class imbalance in the binary classification task, which aligns with the distribution of our dataset. The focal loss we used is defined as:
\[
    L_{\text{FL}}(p_{t}) = \alpha_t \cdot (1 - p_t)^\gamma \cdot \text{BCE}(\hat{y}, y)
\]
where BCE represents Binary Cross Entropy loss, $p_t$ is the model’s predicted probability for the positive class, $\alpha_t$ is the balancing factor, which is tuned to address the class imbalance and $\gamma$ is the focusing parameter, which is used to control the rate at which easy examples are down-weighted. The overall loss for a mini-batch of $N$ samples is:
\[
L = \frac{1}{N} \sum_{i=1}^{N} L_{\text{FL}}(p_{t,i})
\]
Training is end-to-end via standard backpropagation over all learnable modules. The text encoder is kept frozen, as it was already trained on a large corpus of clinical text and to preserve the pretrained representation quality, giving us structured embeddings for the remarks. 

\subsubsection{Inference:}
During inference, a new lung cancer patient’s preoperative clinical features $\mathbf{c}$ and radiological features $\mathbf{r}$ are processed as follows: 
\begin{enumerate} 
\item \textbf{Clinical Summary and Remarks:} A clinical summary $S=f_{\text{summary}}(\mathbf{c})$ is generated and together with $K$ and $P$, is used as an input to LLM producing remarks $M$. 
\item \textbf{Embedding Computation:} Clinical features, radiological features and remarks are embedded using $f_{c}(\cdot)$, $f_{r}(\cdot)$ and $f_{m}(\cdot)$ to produce $E_c$, $E_r$ and $E_m$, respectively. 
\item \textbf{Prediction:} All embeddings are then fused to create $E$, which is then used in final MLP Network to obtain the prediction~$\hat y$.
\end{enumerate}

\subsubsection{Clinician Intervention:}
At inference, the generated remark $M$ is displayed alongside the predicted risk probability. Surgeons may edit or modify the text when the automated explanation does not fully reflect patient nuance or expert judgment. The modified text is re-embedded via $f_m(\cdot)$ and reinjected into the fusion network, updating the predicted risk score in real time. This design allows human expertise to intervene transparently and measurably within the prediction loop, while preserving a structured AI reasoning trail.

\begin{figure*}[h!]
    \centering
    \includegraphics[width=0.8\textwidth]{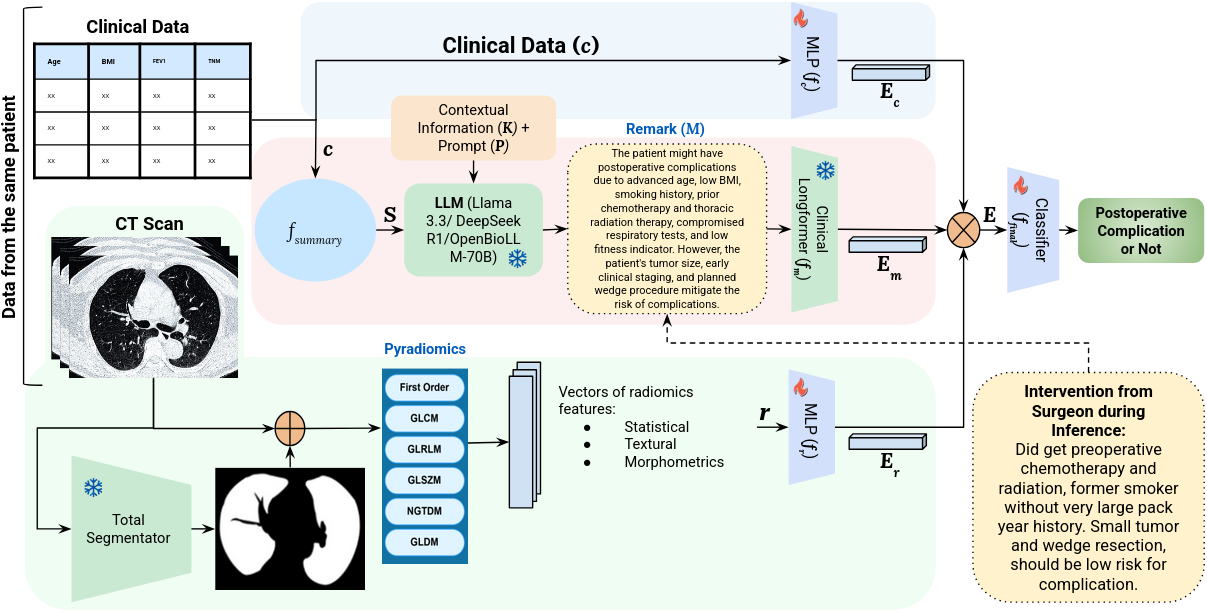}
    \caption{Illustration depicting MIRACLE, our intervenable Deep Learning architecture that combines clinical and radiological features, along with providing surgeons with a point of intervention and explainability. (Best viewed digitally)}
    \label{fig:architecture}
\end{figure*}
\section{Experiments}
\label{sec:experiments}
\subsection{Dataset and Preprocessing}
We conducted this study on \textbf{POC-L}, a real world dataset of 3094 lung cancer patients who underwent surgical resection at Roswell Park Comprehensive Cancer Center between 2009 and 2023. The dataset reflects real clinical cases, preserving the demographic imbalance and clinical heterogeneity, typical of routine thoracic practice. Patients were split into training (2,694; 22.6\% complications), validation (200; 47.5\%) and testing (200; 53.5\%) splits, maintaining the naturally skewed complication distribution. The dataset is dominated with White ethnicity patients with representations from African-American and Asian populations. 57\% of patients are female and 43\% male. Each case has 17 structured preoperative clinical variables. Due to data sharing constraints, preventing transfer of raw CT volumes, image analysis was performed locally prior to feature sharing. Chest CT scans were segmented using TotalSegmentator \cite{Wasserthal_2023}, followed by radiomic feature extraction with PyRadiomics \cite{Pyradiomics}. After refinement and removal of redundant descriptors, 113 standardized radiomic features capturing lung morphology, tumor characteristics and texture statistics were retained. Continuous clinical features were normalized using Min-Max scaling fitted on the training split, while categorical variables were label encoded. Radiomic features were standardized to ensure numerical stability. Postoperative complications were defined across ten major complication events curated by domain experts. For model supervision, the presence of any of the ten complication events was aggregated to produce a binary global outcome label indicating whether a patient experienced at least one postoperative complication. This labeling formulation mirrors real world clinical risk assessment, enabling direct benchmarking of MIRACLE’s predictions against surgical expert judgment under routine care conditions. Due to patient privacy restrictions, POC-L dataset cannot be publicly released. However, all records were de-identified prior to analysis and the study was approved under IRB protocol \textbf{BDR 176423}. For reproducibility, we publicly release our full training pipeline and baselines (see abstract). As per our knowledge, there is no public dataset that matches POC-L’s structure, limiting cross dataset benchmarking. Ongoing efforts are aimed at acquiring and harmonizing multi-institutional datasets to support broader validation and robustness analysis, as POC-L is dominated by one ethnicity, proving as one of the limitation of dataset.

\subsection{Evaluation Metrics}
Traditional measures such as accuracy, precision, recall and F1-score depend heavily on the choice of decision threshold which can be biased during comparison of performance across models. Thus to assess predictive performance and ensure fair comparison with baseline methods, we adopted threshold-independent evaluation metrics. As all models output continuous probabilities , we evaluate performance using AUC and clinically relevant True Positive Rate (Sensitivity) at fixed False Positive Rates (1-Specificity) of 0.2 and 0.3. AUC evaluates discrimination capability across all thresholds, providing a robust, unbiased measure of classification performance. The selected TPR$@$FPR operating points reflect clinically relevant trade-offs, where high sensitivity is prioritized while allowing approximately 70–80\% specificity, as guided by surgical collaborators.

\subsection{LLMs and Remark Generation}
For interpretable risk assessment, we employ three open-source LLMs: Llama 3.3 70B-Instruct\footnote{\url{https://huggingface.co/meta-llama/Llama-3.3-70B-Instruct}} (instruction-based), DeepSeek R1 Distill Qwen-32B\footnote{\url{https://huggingface.co/deepseek-ai/DeepSeek-R1-Distill-Qwen-32B}}\cite{deepseekai2025deepseekr1incentivizingreasoningcapability} (general reasoning) and OpenBioLLM-70B\footnote{\url{https://huggingface.co/aaditya/Llama3-OpenBioLLM-70B}}\cite{OpenBioLLMs} (domain-fine-tuned). Models were selected to represent leading open-source performance within each category while remaining feasible under our available computational resources. For each patient, a composite input $\{S,K,P\}$ is prepared as described in section $3.3$. Each combined input is processed via HuggingFace inference checkpoints of the selected LLM on two 95GB NVIDIA H100 GPUs. Generation parameters are set to \texttt{max\_new\_tokens=2000} and \texttt{temperature=0.9}. The resulting remark \(M\) provides a concise, evidence anchored explanation of why the patient is or is not at elevated risk for postoperative complications. The full remark generation code is provided with our codebase (link in abstract).

\subsection{Baselines}
To contextualize the performance of MIRACLE, we evaluated our approach against two complementary baseline categories: (1) conventional ML classifiers and (2) standalone LLM-based predictors. Additionally, we established a human expert baseline by measuring the performance of practicing surgeons on the held-out test split.

\subsubsection{Classical Machine‐Learning Methods}
We implemented five widely used classifiers: Multivariate Logistic Regression (MVLR), Random Forest Classifier (RFC), XGBoost, Gradient Boosting Classifier (GBC) and LightGBM. All models were trained using combined clinical and radiological features with validation based hyperparameter tuning, following previous medical imaging benchmarks \cite{ExplainableML, noncardiacpostoperative, PBNNs} and thus serve as representative, non deep learning baselines.

\subsubsection{LLM‐Only Predictors}  
As each LLM used in this study adds a particular background and been used for variety of classification task across domains, we evaluated each selected LLM as a standalone reasoning baseline. For this setup, the clinical summary $S$ and knowledge bank $K$ were provided to the LLM alongside a task-specific prompt instructing the model to estimate the probability of postoperative complications. The predicted probabilities were directly used to compute all evaluation metrics. This setup quantifies how much predictive signal is captured by the LLM’s reasoning over structured clinical data using a comprehensive Knowledge Bank. Generation parameters were fixed to \texttt{max\_new\_tokens=500} and \texttt{temperature=0.9}. All inference was performed using HuggingFace checkpoints on two 95GB NVIDIA H100 GPUs.

\subsubsection{Human Baseline}

The objective of our study is to enhance the effectiveness of lung cancer surgery alongside providing a point of intervention and explanation for surgeons. Thus, to benchmark MIRACLE against expert clinical judgment, we evaluated the performance of practicing thoracic surgeons in predicting postoperative complications using the held-out test split. Surgeons from the collaborating cancer research center participated in the study, with clinical experience ranging from 2 to 33 years (mean: 17.75 years). Provided only with preoperative clinical factors and blinded to ground-truth outcomes, surgeons classified each patient as likely or unlikely to develop postoperative complications and submitted brief textual justifications for their decisions, which is something they routinely do in their practice using their expertise and experience. From these annotations, we computed surgeon-level False Positive Rate (FPR) and True Positive Rate (TPR) metrics for direct comparison with our models. Surgeons exhibited an average FPR of approximately 20\%, validating our selection of TPR(\%)$@$FPR=0.2 as a clinically meaningful operating point. Additionally, surgeon-authored remarks were used as qualitative references for evaluating the explanatory quality of the three LLM based remark generation methods, detailed in Section \ref{sec:qualityassessment}.

\noindent All baselines were evaluated on the held‐out test split using AUC and TPR@FPR=0.2/0.3 to ensure direct comparability with our full model. Code for baseline is also released.

\begin{figure*}[htbp]
    \centering
    \includegraphics[height=200px, width=\textwidth]{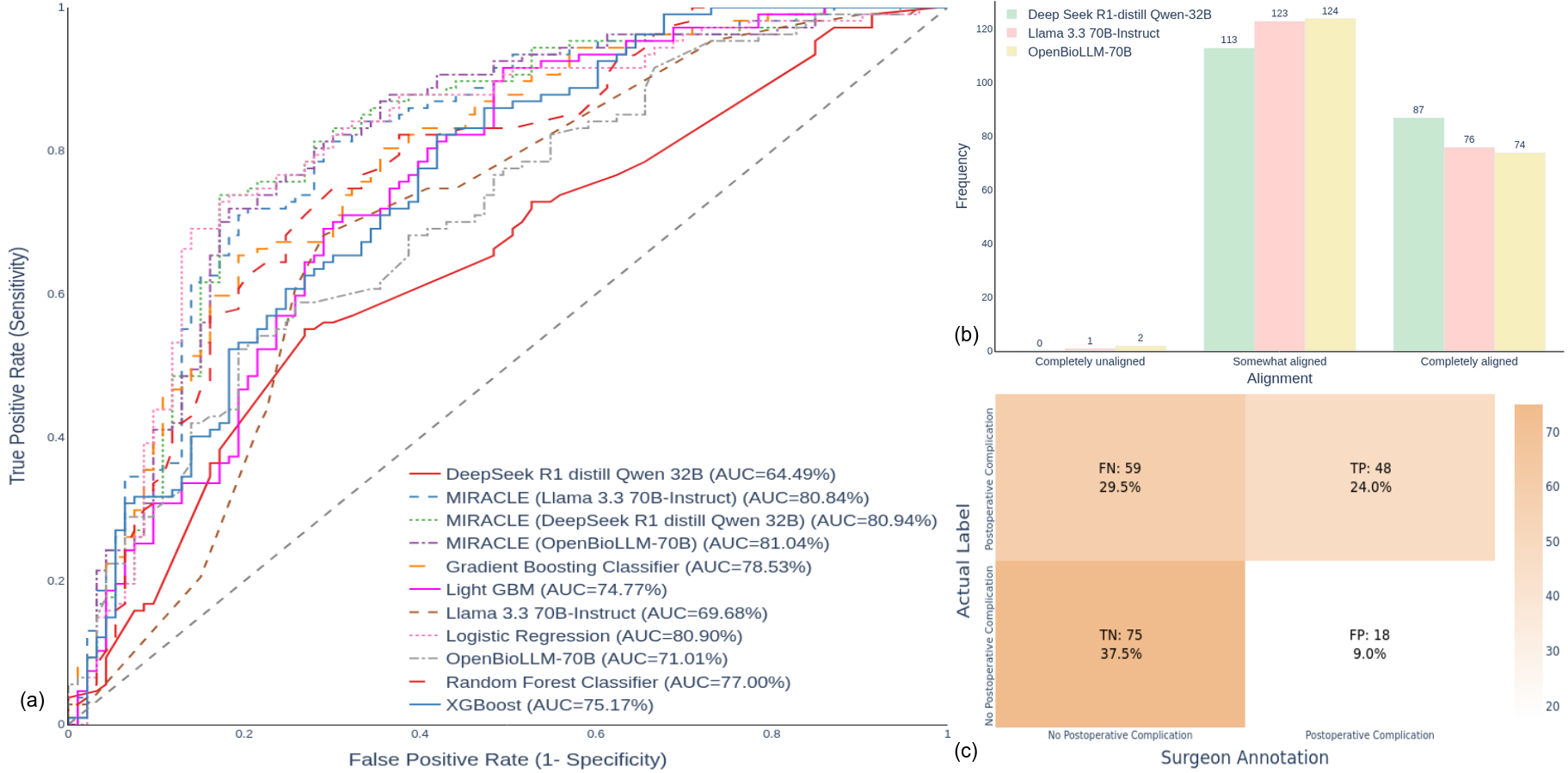}
    \caption{(a) Receiver Operating Curves for MIRACLE alongside all the baseline models. 
    (b) Distribution of alignment between LLM generated remarks and surgeon‐provided remarks.
    (c) Confusion Matrix for the surgeons' performance on POC-L test split.
    (Best Viewed Digitally)}
    \label{fig:figure_4_combined}
\end{figure*}

\subsection{Implementation Details}
\subsubsection{Clinical and Radiological Encoder}
The clinical encoder is a four-layer Bayesian MLP with output dimensions \([64, 128, 256, 768]\). Each layer employs variational Bayesian linear transformations with KL-divergence regularization (\(\lambda = 10^{-6}\)), Monte Carlo sampling using \(S=10\) samples per forward pass and 30\% dropout for regularization. The radiological encoder follows a similar design, implemented as a two-layer Bayesian MLP with output dimensions \([256, 768]\), using the same KL regularization weight, sampling strategy (\(S=10\)) and dropout rate (30\%).

\subsubsection{Remark Encoder}  
For textual encoding, we use Clinical Longformer \cite{li2022clinicallongformerclinicalbigbirdtransformerslong}, a Longformer-based model pretrained on the MIMIC-III clinical notes corpus \cite{mimic}. This model supports sequences of up to 4,096 tokens, enabling full-length encoding of the generated remarks without truncation. The pooled output layer (dimension 768) is passed through a frozen linear projection of identical dimensionality to preserve pretrained clinical representations.

\subsubsection{Feature Fusion}  
Let $\mathbf{E}_c$, $\mathbf{E}_r$ and $\mathbf{E}_m \in \mathbb{R}^{768}$ denote embeddings from the clinical, radiological and remark encoders, respectively. These embeddings are fused via a weighted combination:
\[
\mathbf{E} = 0.5\mathbf{E}_c + 0.25\mathbf{E}_r + 0.25\mathbf{E}_m,
\]
where fusion weights were tuned through extensive experimentation on the validation set.

\subsubsection{Classifier}  
The fused embedding $\mathbf{E}$ is input to a three-layer Bayesian MLP classifier with hidden dimensions \([256, 1024]\) and a single-unit output layer. The same KL regularization (\(\lambda =10^{-6}\)) and Monte Carlo sampling strategy (\(S=10\)) are applied throughout.

\subsubsection{Training}  
The full network is trained using focal loss \cite{lin2018focallossdenseobject} with balancing parameter $\alpha_t=0.8$ and focusing parameter $\gamma=4$ to mitigate class imbalance. Training is performed on 2,694 samples with model selection based on validation AUC computed over a held-out set of 200 patients. Final evaluation is conducted on an independent test cohort of 200 patients. All experiments were executed on a single NVIDIA A6000 GPU, requiring approximately 2GB of VRAM per training run.
\section{Results}
\label{sec:results}
\begin{table}[t]
    \centering
    \scriptsize
    \setlength{\tabcolsep}{3pt}
    \begin{threeparttable}
    \caption{Performance of different models on our testing dataset}
    \label{table:sota}
    \begin{tabularx}{\columnwidth}{>{\raggedright\arraybackslash}X|c|c|c}
        \hline
        Model & \makecell{AUC\\(\%)}
              & \makecell{TPR(\%) \\ @ FPR \\ = 0.2}
              & \makecell{TPR(\%) \\ @ FPR \\ = 0.3} \\  
        \hline
        Llama 3.3 70B-Instruct\tnote{†}  & 69.68  & 41.12 &  74.77 \\
        DeepSeek R1-Distill Qwen-32B\tnote{†}  &  64.49 & 54.21  &  56.07 \\
        OpenBioLLM-70B\tnote{†} \cite{OpenBioLLMs}  & 71.01  & 52.34  &  60.75 \\
        Multivariate logistic regression   & 80.89   & \textbf{73.83} & 80.37   \\
        Random Forest Classifier   & 77.00 & 62.62   & 74.76  \\  
        XGBoost  & 75.17  & 53.27  & 64.48  \\
        Gradient Boosting Classifier  & 78.53  & 65.42 & 67.29  \\
        LightGBM  & 74.77  & 46.73 & 69.16  \\
        Surgeons  & --  & 44.86  & --  \\
        MIRACLE (DeepSeek R1 distill)  & 80.94  & \textbf{73.83}  & \textbf{81.31}  \\
        MIRACLE (Llama 3.3 70B-Instruct)  & 80.84  & 71.03  & \textbf{81.31}  \\
        MIRACLE (OpenBioLLM-70B)  & \textbf{81.04}  & 71.96  & \textbf{81.31} \\
        \hline
    \end{tabularx}
    \begin{tablenotes}
      \item[†] Models using only clinical dataset.
    \end{tablenotes}
  \end{threeparttable}
  \vspace{-1.0em}
\end{table}

\begin{table}[t]
    \centering
    \caption{Ablation Study}
    \label{tab:ablation}
    \setlength{\tabcolsep}{2pt} 
    \renewcommand{\arraystretch}{1.0}
    \scriptsize
    \begin{tabularx}{\columnwidth}{c|c|c|
        >{\centering\arraybackslash}X|
        >{\centering\arraybackslash}X|
        >{\centering\arraybackslash}X}
        \hline
        Clinical & Radiology & \makecell{LLM \\ Remarks}  
                 & \makecell{AUC\\(\%)} 
                 & \makecell{TPR(\%)\\@ FPR \\= 0.2} 
                 & \makecell{TPR(\%)\\@ FPR \\= 0.3} \\
        \hline
        \checkmark & -- & --          & 74.81 & 57.94 & 66.35 \\
        \checkmark & \checkmark & --   & 78.64 & 64.48 & 76.64 \\
        \checkmark & \checkmark & \checkmark & \textbf{80.94} & \textbf{73.83} & \textbf{81.31} \\
        \hline
    \end{tabularx}
\end{table}

\subsection{Quantitative Performance}
Table \ref{table:sota} summarizes the comparative performance of MIRACLE (with all three LLM variants) against traditional machine learning baselines, standalone LLM predictors and the human expert baseline on the held-out test cohort. When integrated with OpenBioLLM-70B \cite{OpenBioLLMs}, MIRACLE achieves the highest overall discrimination with an AUC of \textbf{81.04\%}, closely followed by the DeepSeek R1–Distill Qwen-32B and Llama 3.3 70B–Instruct variants at 80.94\% and 80.84\%, respectively. Corresponding ROC curves are shown in Figure \ref{fig:figure_4_combined}(a). At fixed operating points, all MIRACLE variants attain the highest sensitivity at FPR=0.3 with a TPR of \textbf{81.31\%}, outperforming all traditional baselines. At the stricter threshold of FPR=0.2, MIRACLE with the DeepSeek R1–Distill variant matches the strongest classical baseline (multivariate logistic regression) with a TPR of \textbf{73.83\%}, while the OpenBioLLM-70B and Llama 3.3 variants remain highly competitive at 71.96\% and 71.03\%, respectively. Although multivariate logistic regression demonstrates comparable sensitivity at the lowest false positive rate, it lacks any inherent interpretability or intervention capability. In contrast, MIRACLE combines state of the art predictive performance with feature grounded, clinician editable explanations, providing actionable insights alongside probability estimates. Performance differences among MIRACLE variants are marginal, indicating stable multimodal fusion irrespective of the underlying LLM. OpenBioLLM-70B yields slightly superior overall discrimination (highest AUC), whereas DeepSeek R1–Distill exhibits the strongest sensitivity at strict operating points, supporting greater early risk detection which is an important consideration for surgical planning. Standalone LLM predictors show noticeably lower performance but their generated summaries provide complementary semantic context that strengthens multimodal fusion within MIRACLE. Notably, relative performance trends observed in standalone LLM testing remain consistent once integrated into the full system. Human experts, with an average experience of 17.75 years, achieved a TPR of 44.86\% at FPR=0.2, exceeding the standalone Llama baseline but falling substantially below MIRACLE’s performance. This gap highlights the added clinical value of our intervention-enabled, multimodal risk stratification framework. Surgeon confusion matrices are shown in Figure \ref{fig:figure_4_combined}(c).

\subsection{Qualitative Analysis of the generated remarks}
\label{sec:qualityassessment}
To assess the quality of LLM-generated clinical explanations, we evaluated all selected LLMs using a two stage qualitative framework. Despite identical inputs, each model produced distinct remarks, motivating a systematic comparison against surgeon-authored explanations.

 {
\setlength{\parskip}{0pt}
  \textbf{(i) Automated Adjudication.}  
    Following the LLM as a judge protocol \cite{zheng2023judgingllmasajudgemtbenchchatbot}, we used DeepSeek R1-Distill Qwen-7B \cite{deepseekai2025deepseekr1incentivizingreasoningcapability} to evaluate semantic alignment between each LLM’s remarks and corresponding surgeon authored explanations. For each pair, alignment was categorized as \emph{completely unaligned}, \emph{somewhat aligned} or \emph{completely aligned}. As shown in Figure \ref{fig:figure_4_combined}(b), nearly all outputs were judged aligned to that of surgeons', indicating strong grounding and limited requirement for intervention. Frequency analysis showed that the reasoning based DeepSeek R1-Distill Qwen-32B achieved the highest complete alignment rate (43.5\%), followed by the instruction-based Llama 3.3 70B-Instruct (38\%). OpenBioLLM-70B \cite{OpenBioLLMs} yielded the lowest fully aligned rate (37\%) but demonstrated the highest proportion of partially aligned responses, suggesting broader hypothesis exploration.
    
  \textbf{(ii) Expert Manual Review.}  
    A panel of thoracic surgeons conducted manual assessments of paired surgeon and LLM explanations in the test split. Each LLM remark was categorized as:
    (i) \emph{Performs better} when it is clearer or more clinically informative than surgeons' remarks,
    (ii) \emph{Performs comparably} when it captures similar insights and key risk factors, or
    (iii) \emph{Performs worse} when it omitted critical risk factors, made implausible inferences, is too verbose or overestimates the risks.
    While many LLM explanations met or exceeded surgeon clarity, experts confirmed instances where surgeon insights remained superior. These findings validate our design choice to incorporate a clinician intervention loop, enabling surgeons to revise generated explanations and immediately update predicted risk scores. Representative examples from the \emph{Performs better} and \emph{Performs worse} categories are provided in a figure in supplementary material, with additional examples across all three categories also provided.
This hybrid qualitative evaluation framework enables reproducible comparison of explanation quality across all LLM variants.
}

\subsection{Ablation Study}
\label{sec:ablation}
We conducted ablation study using the strongest MIRACLE configuration, DeepSeek R1–Distill Qwen-32B variant, evaluated on the held-out test split of 200 patients as it outperformed all the baselines across all metrics (Table \ref{tab:ablation}). Each ablation isolates the contribution of individual architectural components. Using only the clinical Bayesian MLP and classifier yields an AUC of 74.81 with reduced TPR at both operating thresholds, surpassing standalone LLM baselines but underperforming conventional ML models (Table \ref{table:sota}). Adding the radiological encoder increases AUC by approximately 4\%, demonstrating the complementary value of radiomic biomarkers. This dual encoder configuration exceeds nearly all baselines, improving TPR by roughly 6.5\% at FPR=0.2 and 10\% at FPR=0.3. Incorporating the LLM remark encoder provides a further performance gain, increasing AUC by 2.3\%, raising TPR@FPR=0.2 by 9.35\% and improving TPR@FPR=0.3 by approximately 5\%. While remark embeddings contribute modestly to overall discrimination, they substantially enhance sensitivity under strict clinical operating points and introduce interpretable explanations that enable surgeon intervention, supporting earlier identification of high-risk surgical candidates.
\section{Discussion}
\label{sec:discussion}
MIRACLE addresses a central clinical challenge in thoracic oncology which is accurate preoperative risk stratification for postoperative pulmonary complications. Current standards of care rely primarily on population level pulmonary function thresholds such as FEV1 and DLCO, along with subjective clinician judgment, which fail to capture patient specific heterogeneity in lung structure and comorbidity interactions. Our results show that incorporating CT derived radiomic biomarkers significantly improves discrimination and sensitivity, demonstrating that imaging encodes latent markers of pulmonary risk not reflected in conventional metrics. By integrating clinical variables, radiomics and language based reasoning, MIRACLE aligns computational prediction with how surgeons assess risk in practice, through multimodal interpretation of physiology and treatment context.

Apart from predictive performance, MIRACLE’s primary advancement lies in transforming risk assessment from a static probability output into an interactive decision support workflow. LLM generated explanations expose the reasoning behind predictions and enable clinicians to directly revise textual remarks, which are then again used to update risk estimates. Experts' evaluation confirmed value of this human in the loop paradigm, particularly in cases where surgeon insight exceeded automated explanations. In contrast to previous explainability approaches that merely visualize feature importance, MIRACLE enables direct clinician intervention within the predictive process, preserving accountability while leveraging AI support. 

Clinically, MIRACLE has the potential to support individualized peri-operative planning including targeted prehabilitation programs, ventilatory optimization strategies, selective ICU triage and clearer informed consent based on patient specific risk factors. MIRACLE integrates into existing preoperative workflows via a web based interface (screenshot in supplementary material), requires average computational resources and can operate using extracted radiomic features without requiring direct access to imaging data, facilitating deployment even in small clinics or resource constrained hospitals where image transfer is restricted.

Before routine clinical use, several validation steps remain essential. The current training population is predominantly White and single-institutional. External, multi-institutional datasets are required to ensure generalizability and mitigate demographic bias. Local finetuning may be necessary to align performance with regional demographics and care pathways. Domain shifts related to CT acquisition protocols and reconstruction parameters also pose risks to robustness. Additionally, although anchored to curated knowledge sources, LLM generated remarks retain potential for factual inconsistency and bias. Common failure modes observed include omission of complex comorbidity interactions and overestimation of pulmonary risk. Importantly, MIRACLE is designed to support and not replace clinical judgment, with surgeons retaining final decision authority. Ongoing multi institutional data acquisition and prospective validation studies will be important to confirm safety, fairness and regulatory readiness. Collectively, MIRACLE represents a clinically aligned framework that unifies prediction, explanation and actionable intervention capabilities, which is not offered by existing risk calculators. We plan to extend this framework by using images directly, making the process of prediction more dynamic and extending MIRACLE to other types of cancers.
\section{Conclusion}
\label{sec:conclusion}
We presented MIRACLE, a unified framework to integrate preoperative clinical features, CT derived radiomics and LLM generated natural language explanations for postoperative complication risk prediction in lung cancer surgery. Across extensive evaluation on a real world dataset, MIRACLE demonstrated consistent improvements over all baselines (ML models, LLMs, experts). MIRACLE also introduces a human in the loop paradigm, which enables transparent explanation and direct clinician intervention, transforming risk prediction from a static output into an interactive decision process. However, external testing across institutions with diverse imaging protocols and prospective evaluation of decision impact will be necessary to ensure generalizability and safety. By combining performance, interpretability, intervenability and practical deployability, MIRACLE offers a scalable approach to improve peri-operative planning and patient counseling and highlights the potential of integrated AI to deliver safer and more personalized surgical care.
{
    \small
    \bibliographystyle{ieeenat_fullname}
    \bibliography{main}
}

\end{document}